%% file: main.tex
\documentclass[letterpaper]{article} 
\usepackage{aaai25}  
\usepackage{times}  
\usepackage{helvet}  
\usepackage{courier}  
\usepackage[hyphens]{url}  
\usepackage{graphicx} 
\usepackage{natbib}  
\usepackage{caption} 
\usepackage{algorithm}
\usepackage{algorithmic}
\usepackage{amsmath}
\usepackage{amssymb}
\usepackage{multirow}
\DeclareMathOperator*{\argmax}{arg\,max}

\usepackage{color, colortbl}
\definecolor{Gray}{rgb}{0.92,0.92,0.92}
\usepackage{newfloat}
\usepackage{listings}
\DeclareCaptionStyle{ruled}{labelfont=normalfont,labelsep=colon,strut=off} 
\floatstyle{ruled}
\newfloat{listing}{tb}{lst}{}
\floatname{listing}{Listing}
\title{Multi-Scale Contrastive Learning for Video Temporal Grounding}
\author{
    Thong Thanh Nguyen\textsuperscript{\rm 1}, \;
    Yi Bin\textsuperscript{\rm 1,2\thanks{Yi Bin is the corresponding author, yi.bin@hotmail.com}}, \;
    Xiaobao Wu\textsuperscript{\rm 3}, \;
    Zhiyuan Hu\textsuperscript{\rm 1}, \; \\
    Cong-Duy Nguyen\textsuperscript{\rm 3}, \;
    See-Kiong Ng\textsuperscript{\rm 1}, \;
    Anh Tuan Luu\textsuperscript{\rm 3}
}
\affiliations{
    \textsuperscript{\rm 1} Institute of Data Science (IDS), National University of Singapore, Singapore \\
    \textsuperscript{\rm 2} Tongji University, China,
    \textsuperscript{\rm 3} Nanyang Technological University (NTU), Singapore \\

}

\usepackage{bibentry}
\usepackage{color}
\definecolor{orange}{RGB}{253,107,9}

\newcommand{\answerTODO}[1][]{\textcolor{red}{\bf [TODO]}}

\begin{document}

\maketitle

\begin{abstract}
Temporal grounding, which localizes video moments related to a natural language query, is a core problem of vision-language learning and video understanding. To encode video moments of varying lengths, recent methods employ a multi-level structure known as a feature pyramid. In this structure, lower levels concentrate on short-range video moments, while higher levels address long-range moments. Because higher levels experience downsampling to accommodate increasing moment length, their capacity to capture information is reduced and consequently leads to degraded information in moment representations. To resolve this problem, we propose a contrastive learning framework to capture salient semantics among video moments. Our key methodology is to leverage samples from the feature space emanating from multiple stages of the video encoder itself requiring neither data augmentation nor online memory banks to obtain positive and negative samples. To enable such an extension, we introduce a sampling process to draw multiple video moments corresponding to a common query. Subsequently, by utilizing these moments' representations across video encoder layers, we instantiate a novel form of multi-scale and cross-scale contrastive learning that links local short-range video moments with global long-range video moments. Extensive experiments demonstrate the effectiveness of our framework for not only long-form but also short-form video grounding.
\end{abstract}

%

\input{sections/01_introduction}
\input{sections/02_related_work}
\input{sections/03_methodology}
\input{sections/04_experimental_setup}

\input{sections/05_experimental_results}
\input{sections/06_conclusion}
\input{sections/07_acknowledgement}

\bibliography{aaai25}

\end{document}

%% file: sections/01_introduction.tex
\section{Introduction}
Temporal video grounding aims to localize moments of interest in an untrimmed video given a free-form textual description. It is a challenging multimodal task since it involves understanding temporal information in videos and reasoning about their connections to semantic information in texts. Recently, temporal grounding has drawn increasing attention \citep{mu2024snag, jung2023overcoming, xu2023boundary, pan2023scanning}, due to its wide range of applications such as surveillance \citep{zhang2016context}, robotics \citep{burgner2015continuum}, and autonomous driving \citep{claussmann2019review}.

Previous methods \citep{zhang2020learning, soldan2021vlg, zhang2020span} for temporal grounding concentrate on grounding merely a few queries in short video snippets. However, recently the growing availability of long videos, \textit{e.g.} on streaming platforms, and demands to query their rich content have necessitated productive grounding of large volumes of queries in long videos. Because of such short-to-long video paradigm shift, latest methods \citep{zhang2022actionformer, mu2024snag} have utilized local self-attention to restrict attention within a local window, following the intuition that temporal context beyond a certain range is less helpful for moment localization.

\begin{figure*}[t]
    \centering
    \includegraphics[width=0.19\linewidth]{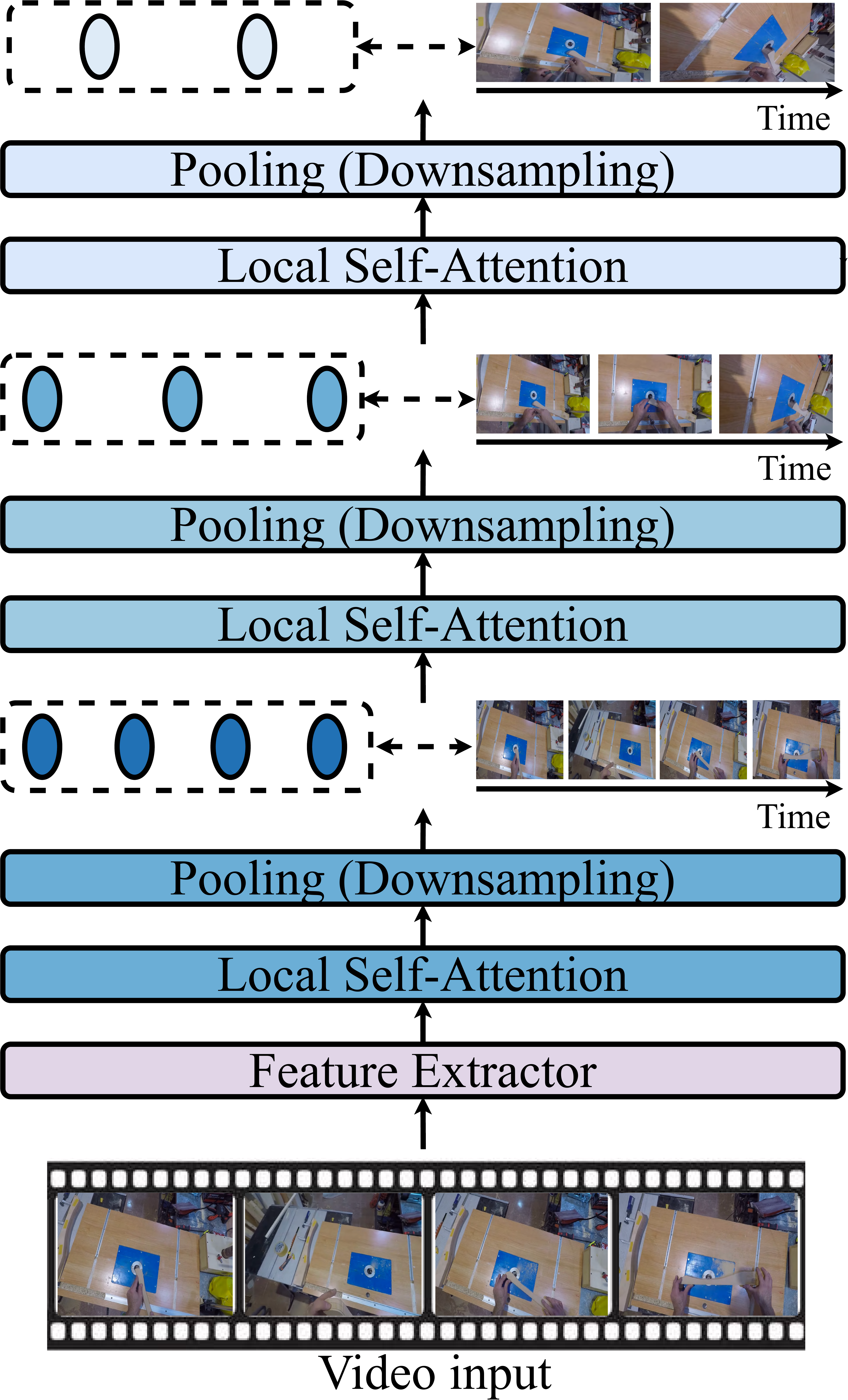} \;\;
    \includegraphics[width=0.77\linewidth]{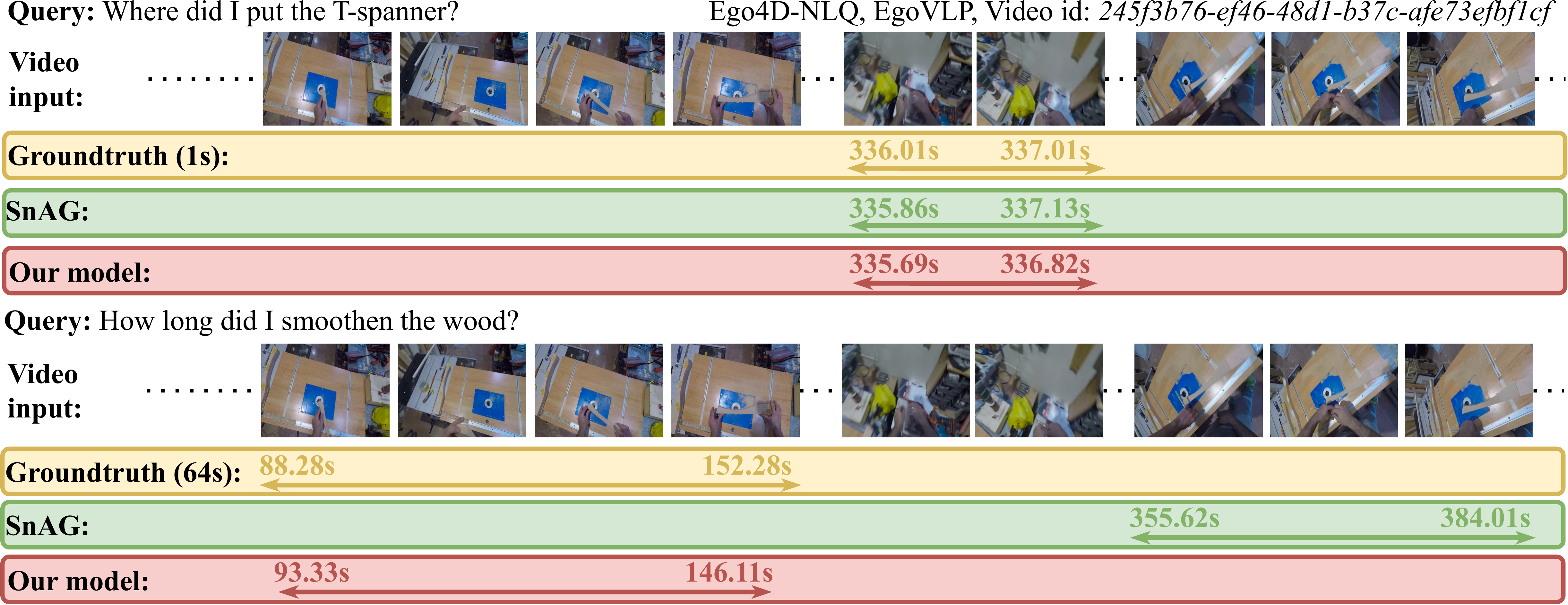} 
    \caption{(Left) Illustration of feature pyramid to encode video moments of different lengths; (Right) An example where recent method SnAG \citep{mu2024snag} accurately localizes short video moment but fails on long moment. }
    \label{fig:temporal_grounding_example}
\end{figure*}

\begin{figure*}[t]
\centering
\includegraphics[width=0.24\linewidth]{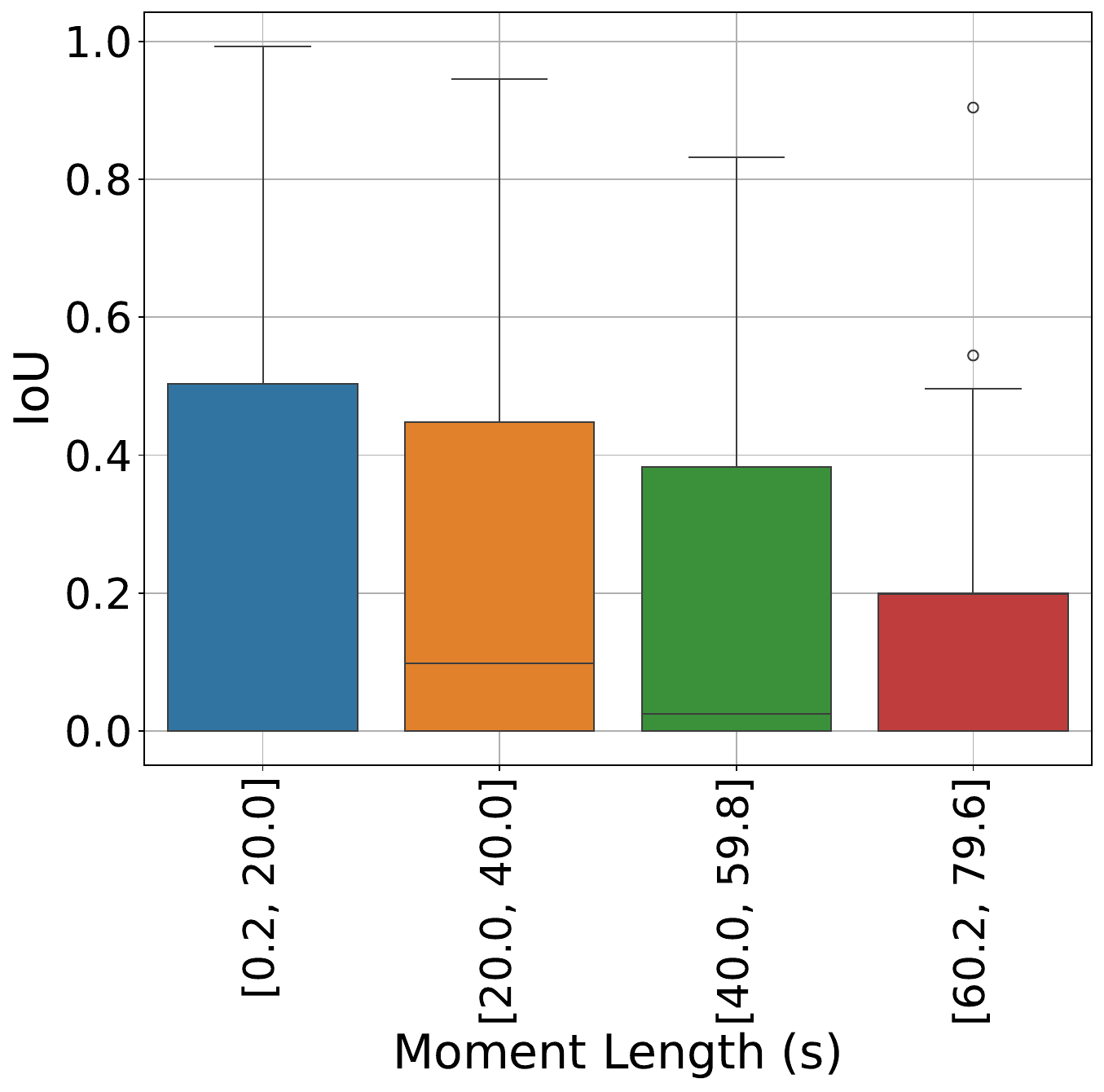}\,
\includegraphics[width=0.24\linewidth]{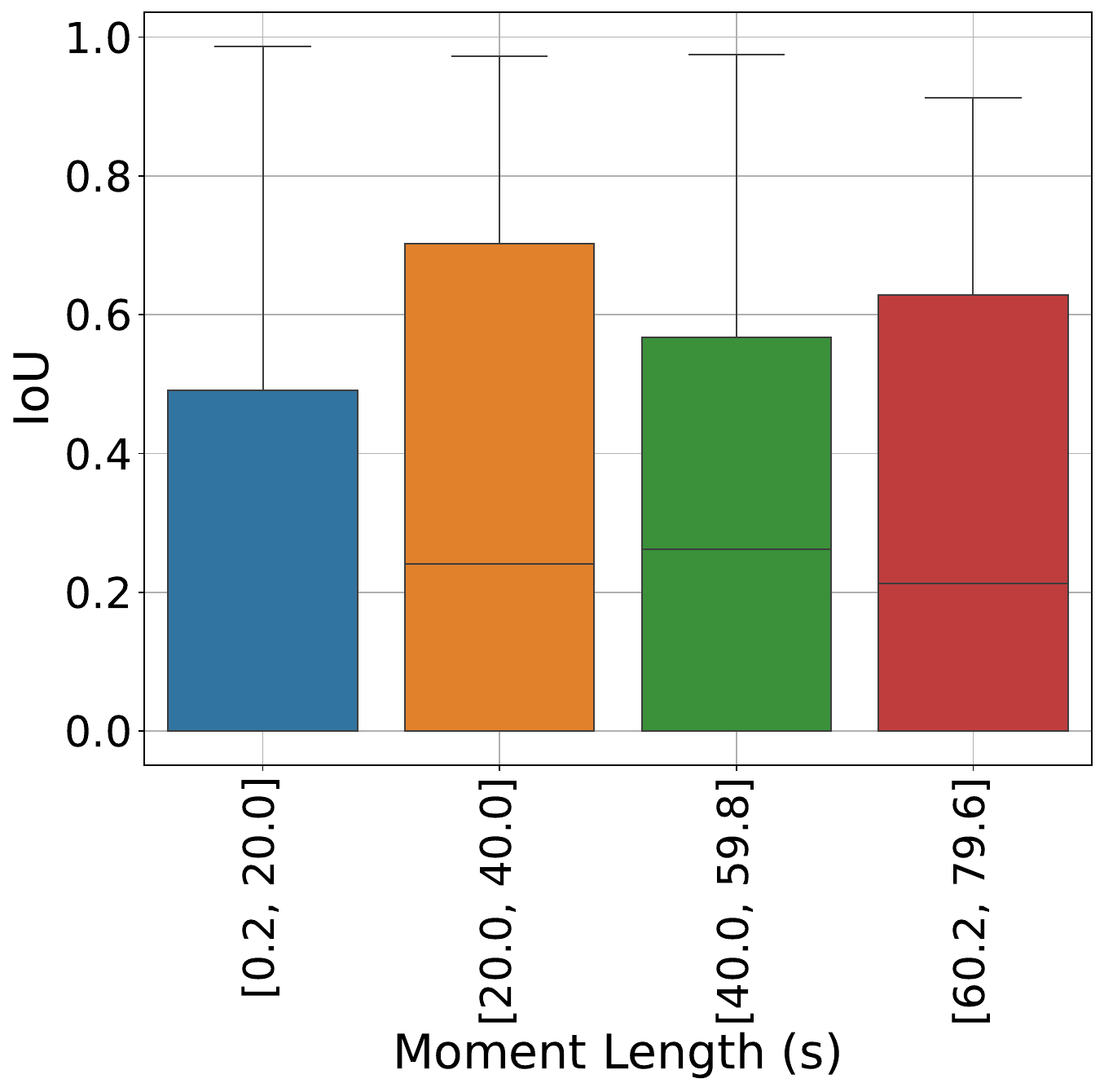}
\includegraphics[width=0.24\linewidth]{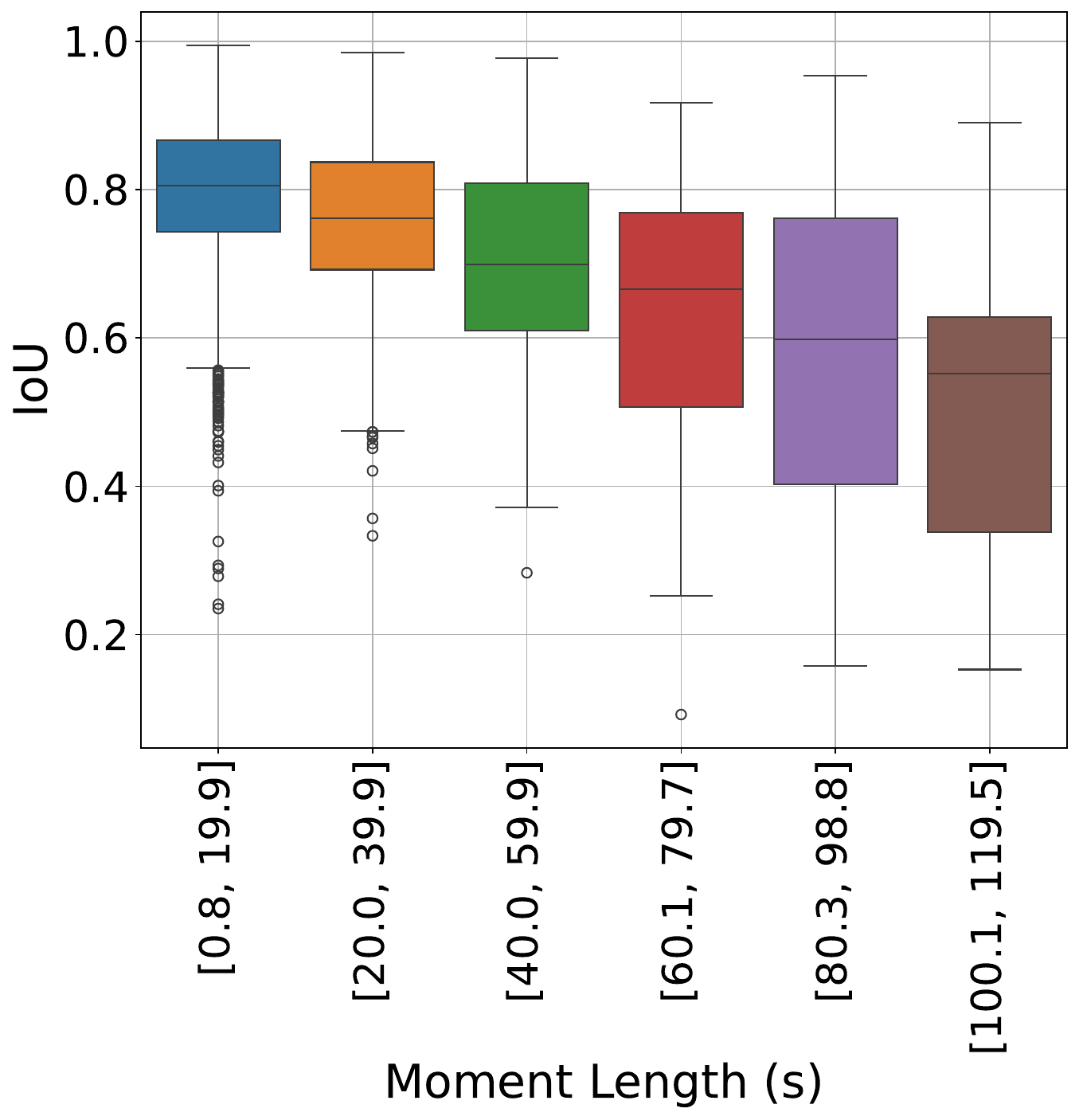}
\includegraphics[width=0.24\linewidth]{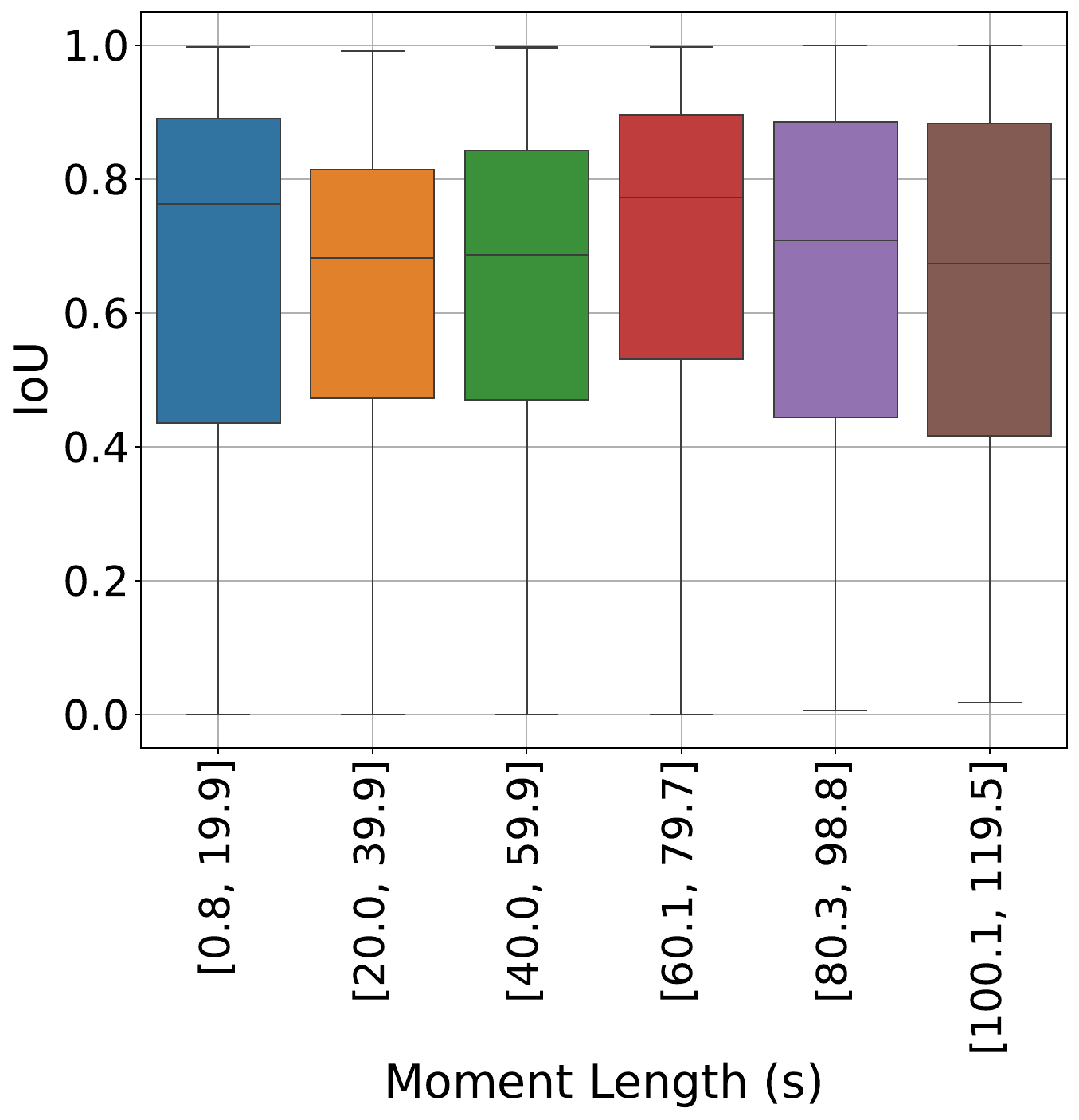}
\caption{First and Second: IoU results with respect to target video moment length on Ego4D-NLQ \citep{grauman2022ego4d} of baseline SnAG \citep{mu2024snag} and our model. Third and Fourth: IoU results with respect to target video moment length on TACoS \citep{regneri2013grounding} datasets of baseline SnAG \citep{mu2024snag} and our model.}
\label{fig:ego4d_tacos_moment_length_iou}
\end{figure*}

To capture moments at different temporal scales without enlarging the window size of the local self-attention, recent methods \citep{zhang2022actionformer, mu2024snag} need to combine several Transformer blocks with downsampling between every two blocks, resulting in a feature pyramid of moment representations, as illustrated in Figure \ref{fig:temporal_grounding_example} (left). Unfortunately, due to such downsampling operation, when moment representations are propagated from lower levels of short-range (local) moments to higher levels of long-range (global) moments, information contained in representations of longer moments will gradually degrade \cite{guo2020augfpn, yang2023afpn}. This could explain why performance of these methods tends to degrade as the duration of target moments increase, as shown in Figure \ref{fig:temporal_grounding_example} (right) and statistically shown with Intersection-over-Union (IoU) results in Figure \ref{fig:ego4d_tacos_moment_length_iou}, respectively.

To enrich information in video moment representations, recent works \citep{panta2024cross, xiao2024bridging, ji2024weakly, liu2024towards} have employed contrastive learning for temporal grounding. The intuition is to capture mutual information between video moments and textual query to preserve salient semantics in moment representations. These works mainly involve query-moment pairs in which queries relate to video moments of distinct videos, hence the learned semantics among moment representations would be independent from each other. However, such approach might not be suitable for the latest scalable video-centric approach \citep{zhang2022actionformer, mu2024snag}, in which multiple textual queries are related to one video. Therefore, if the grounding of two textual queries results in temporal overlapping, there might be a conflict in compact moment representations \citep{an2023unicom}. Furthermore, focusing upon moment-query relations limits these works to the feature space of the final encoder layer, which could not effectively utilize all hidden representations across encoding layers. For multi-scale temporal grounding, such cross-scale representations should be fully used since they express semantics in video moments of various lengths. 

To resolve the above issues, in this paper, we propose a multi-scale contrastive learning framework for multi-scale temporal grounding. In our framework, instead of leveraging moment-query relationships, we utilize the association among video moments. Particularly, to avoid representation conflict among video moments, we introduce a query-centric contrastive approach that draws temporally separate video moments corresponding to a common textual query. A central component of our framework is the creation of positive and negative video moment samples, which previous works primarily apply data augmentation \citep{kim2022exploring, xing2023svformer}. However, because most long-form videos consist of a high volume of video moments, choosing an appropriate augmentation strategy that suits every moment is a non-trivial and lengthy tuning step.  Another common approach is to introduce a memory bank to store positive or negative samples' representations, which are created by aggregating input representations iteratively during training \citep{panta2024cross, han2023momentum}. Nevertheless, a memory bank would present additional hyperparameters such as the bank size and update frequency, which demand laborious tuning effort \citep{wang2021exploring}. 

To prevent these problems, we directly draw samples from the feature space of video moment encoder. Specifically, we take advantage of internal, intermediate representations of video moments from the encoder that are readily available through the feed-forward step of the network without the need to rely upon external steps such as data augmentation or online storing of samples in memory banks. Accordingly, we introduce a within-scale and cross-scale approach to create positive and negative moment samples for contrastive learning. Regarding the within-scale approach, we seek to pull together representations of such semantically close video moments on the same scale of similar temporal range. Moreover, we also push apart representations of video moments which are unrelated to the textual query. Regarding the cross-scale approach, we compel the model to relate global long-range video moments to local short-range moments, while simultaneously repelling semantically distant cross-scale representations in an analogous cross-scale manner. This cross-scale approach would enable long-range moment representations to capture nuanced details of short-range moments, thereby mitigating informational degradation within long-range representations.

To sum up, our contributions are the following:
\begin{itemize}
    \item We propose a multi-scale contrastive framework that focuses on moment-moment relations to mitigate informational degradation in video moment representations.
    \item We propose a within- and cross-scale strategy that supports semantic consistency not only between similar-range but also cross-range video moment representations emanating across layers of the video encoder.
    \item Our framework achieves superior results across major benchmark datasets concerning both short-form and long-form video grounding.
\end{itemize}

%% file: sections/02_related_work.tex
\section{Related Work}
\noindent\textbf{Temporal Grounding.} Research works on temporal grounding can be categorized into two groups: two-stage and single-stage. In the two-stage group, methods generate temporal segments as proposals, then score the segments' probabilities of being target moments and predict the refined boundary timestamps. Early approaches \citep{anne2017localizing, gao2017tall} densely sample proposals leveraging sliding windows and score the proposals independently. Instead of independent sampling, \citet{liu2021adaptive, xiao2021natural, xiao2021boundary} subsequently condition proposal generation upon sentence queries and/or video context to avoid dense sampling. In contrast, \citet{gao2021relation, soldan2021vlg, wang2021structured, zhang2021multi} enumerate all segments and organize them into a 2D adjacency map for relation prediction. In the single-stage group, methods localize moments in a single shot without utilizing proposals, thus being more efficient than the two-stage group. Several works decode moment boundaries from a pooled representation \citep{li2022compositional, li2021proposal, zhou2021embracing} or learnable queries \citep{nguyen2023demaformer, lin2023univtg, woo2022explore}. Most related to our work are models that denote moment candidates as points \citep{fang2023you, liu2022memory, mu2024snag, li2021proposal}. To capture temporal moments of various lengths, these methods designate a feature pyramid to produce moment representations of various scales, resulting in semantic gaps for longer-length moments. 

\noindent\textbf{Contrastive Learning.} Notable improvements contributed by contrastive learning are made by contrastive loss applied to the final encoded outputs \citep{nguyen2021contrastive,wu2023infoctm,nguyen2025meta, hu2021region, nguyen2022adaptive, wu2024modeling,nguyen2024topic, wang2021exploring, nguyen2023improving,nguyen2024kdmcse}. \citet{wang2021exploring, hu2021region} employ a memory bank to maintain an extended set of positive and negative samples. Instead of utilizing outputs at the single final layer, contrastive learning with local and global representations across different layers has been widely studied \citep{hjelm2018learning, zhang2020unsupervised, bachman2019learning, chaitanya2020contrastive}. \citet{zhang2020unsupervised} maximize the mutual information between representations of different local windows of a sentence and the representation of the global sentence. In addition, \citet{bachman2019learning} propose an InfoNCE loss which optimizes over local and global features of two augmented views of an image, whereas \citet{chaitanya2020contrastive} specifically concentrate on medical images. 

%% file: sections/03_methodology.tex
\section{Methodology}
In this section, we delineate our proposed contrastive framework for multi-scale temporal grounding, particularly focusing on a sampling procedure to draw video moment representations across temporal scales.
\begin{figure*}[t]
\centering
\includegraphics[width=0.7\linewidth]{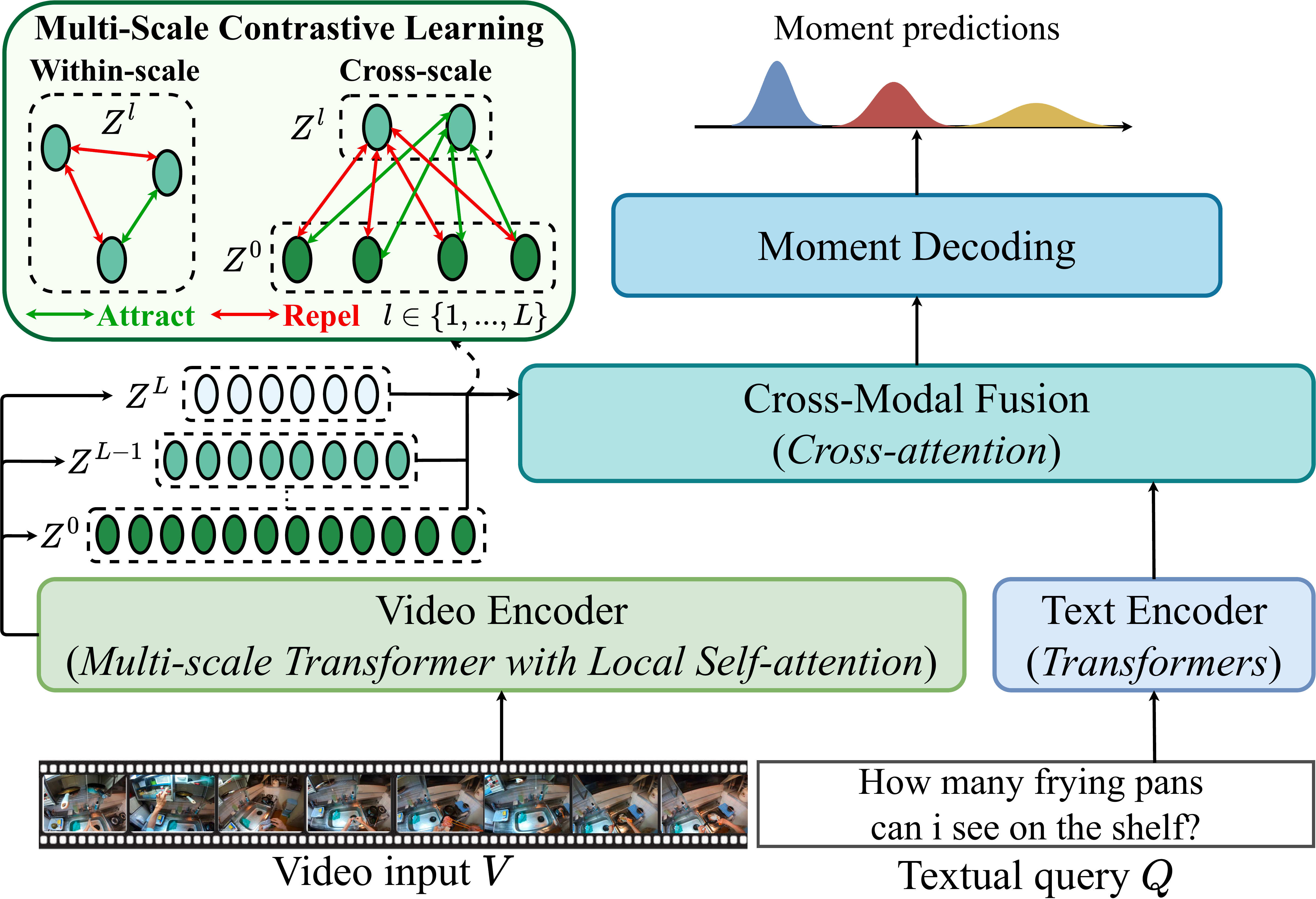}
\caption{Overall illustration of the proposed framework.}
\label{fig:overall_illustration}
\vspace{-10pt}
\end{figure*}

\subsection{Preliminary - Video Temporal Grounding}
We denote an input video $V$ as a sequence of video clips $\{v_{t}\}_{t=1}^{T} = \{v_{1}, v_{2}, ..., v_{T}\}$, where $v_{t}$ denotes a video moment (clip) centered at time $t$. We use a pre-trained feature extractor to embed each  $v_{t}$ into a moment embedding $\mathbf{v}_{t}$. Given the video $V$, our task is to localize a moment $y = \left(s, e\right)$ based on a sentence query $Q = \{q_{1}, q_{2}, ..., q_{K}\}$. Similar to the input video, we also embed the query $Q$ into a sequence of word embeddings $\{\mathbf{q}_{1}, \mathbf{q}_{2}, ..., \mathbf{q}_{K}\}$.

\paragraph{Video encoder.} After embedding video clips, we use a convolution-based projection function to encode local context of video clips:
\begin{equation}
Z^{0} = \{\mathbf{z}_{t}^{0}\}_{t=1}^{T} = \text{Conv}\left(\mathbf{v}_{1}, \mathbf{v}_{2}, ..., \mathbf{v}_{T}\right).
\end{equation}
Subsequently, we designate $L$ Transformer layers to encode temporal context among video clips. In detail, each Transformer layer consists of a local multi-head self-attention (LocalMSA) with a window size of $W$ and an MLP block, in which we restrict the attention to be within a local window:
\begin{gather}
\hspace{-5pt}
\bar{Z}^{l} = \alpha^{l}\cdot\text{LocalMSA} \left(\text{LN}\left(Z^{l-1}\right)\right) + Z^{l-1}, \\
\hat{Z}^{l} = \bar{\alpha}^{l} \cdot\text{MLP}\left(\text{LN}\left(\bar{Z}^{l}\right)\right) + \bar{Z}^{l}, \\
Z^{l} = \downarrow\left(\hat{Z}^{l}\right), \quad l \in \{1, 2, ..., L\},
\end{gather}
where $Z^{l-1}, \bar{Z}^{l}, \hat{Z}^{l} \in \mathbb{R}^{T^{l-1} \times D}$, $Z^{l} \in \mathbb{R}^{T^{l} \times D}$. $T^{l-1} / T^{l}$ is the downsampling ratio, $\alpha^{l}$ and $\bar{\alpha}^{l}$ are learnable per-channel scaling factors \citep{touvron2021going}, $D$ is the hidden dimension, and LN is the layer normalization. 

Inspired by \citep{mu2024snag}, we implement the downsampling operator $\downarrow$ as a strided depthwise 1D convolution. The downsampling operation engenders the multi-scale property of the encoder, generating representations for longer video moments.

\paragraph{Text encoder.} We use Transformer layers, where each layer includes a vanilla self-attention followed by an MLP. Thus, the textual encoder produces textual representations $E = \{\mathbf{e}_{1}, \mathbf{e}_{2}, ..., \mathbf{e}_{K}\}$ for query embeddings $\{\mathbf{q}_{1}, \mathbf{q}_{2}, ..., \mathbf{q}_{K}\}$.

\paragraph{Cross-modal fusion.} Our architecture uses cross-attention to fuse video clip and query word representations. Technically, we modulate video clip representations $\{Z^{l}\}_{l=1}^{L}$ with word representations $E$ as follows:
\begin{gather}
\label{eq:cross_modal_fusion_ln}
\tilde{Z}^{l} = \text{LN}\left(Z^{l}\right), \;\; \tilde{E} = \text{LN}\left(E\right), \\
O^{l} = \sigma\left(\frac{\left(\tilde{Z}^{l}\right)^{\top} \cdot \tilde{E}}{\sqrt{D}}\right) \cdot \tilde{Z}^{l}, \\
X^{l} = \beta^{l} \cdot \text{MLP}\left(\text{LN}\left(O^{l}\right)\right) + O^{l}, 
\end{gather}
where $\beta^{l}$ denotes a learnable per-channel scale and $\sigma$ the Softmax activation function. 

\paragraph{Moment decoding.} After cross-modal fusion, our model converts each time step $t$ to a moment candidate. Specifically, given $\mathbf{x}_{t}^{l}$, we use a convolutional network comprising 1D convolutional layers as the classification head to predict a score $p_{t}^{l}$. In a similar vein, we use a similar 1D convolutional network attached with a ReLU activation function to regress the normalized distances from $t$ to the moment boundaries $(d_{t}^{s}, d_{t}^{e})$ if $\mathbf{x}_{t}^{l}$ is classified as positive. Formally, the decoded moment is computed as:
\begin{gather}
(t, l) = \argmax_{t,l} p_{t}^{l}, \\
\hat{s} = 2^{l-1} \left(t - d_{t}^{s}\right), \quad \hat{e} = 2^{l-1} \left(t + d_{t}^{e}\right).
\end{gather}
During testing, we employ Soft-NMS \citep{bodla2017soft} to merge overlapping moment predictions.

\subsection{Cross-scale Contrastive Learning}
\noindent\textbf{Query-centric sampling.} As randomly sampling moment-query pairs for contrastive learning might lead the model to representation conflict if the groundings of two queries overlap with each other, we instead introduce a sampling approach that draws a text query $Q$ and its temporally separate video moments associated with a common video $V$:
\begin{equation}
Q_{j'}, \{y_{j'}^{l}\}_{l=1}^{L} \sim \mathcal{U} \left(\{Q_{j}, \{y_{j'}^{l}\}_{l=1}^{L}\}_{j=1}^{N_{Q}}\right),
\end{equation}
where $\mathcal{U}$ denotes a discrete uniform distribution, $\{y_{j'}^{l}\}_{l=1}^{L}$ the set of target video moments in each layer $l$, and $N_Q$  the number of textual queries related to video $V$. We generate the target set $\mathcal{P}(l)$ via center sampling \citep{zhang2022actionformer, mu2024snag}, \textit{i.e.} given any moment centered at $t$, any time step $c \in \left[t - \alpha \frac{T}{T^{l}}, t + \alpha \frac{T}{T^{l}}\alpha\right]$  in layer $l$ is considered as a target. After sampling the query and target moments, we directly utilize the representations $\{\mathbf{z}_{j'}^{l}\}_{l=1}^{L}$ of the target moments $\{y_{j'}^{l}\}_{l=1}^{L}$ extracted by the aforementioned multi-scale video encoder.

\paragraph{Within-scale contrastive learning.} Having obtained the representations of target moment samples, we directly utilize moments within each scale as positive and negative samples. Particularly, we iterate over every layer $l$ of the video encoder, and for each anchor video moment $y_{j'}^{l}$, we consider all video moments of layer $l$ corresponding to query $Q_{j'}$ to become positive moment set $P(l)$, and randomly draw those not corresponding to query $Q_{j'}$ to be negative set $\mathcal{N}(l)$. 

Then, we formulate multi-scale contrastive objective over all layers $l \in \{1, 2, ..., L\}$, which pushes positive moments closer while negative moments further:
\begin{equation}
\begin{split}
\mathcal{L}_{\text{within}} &= -\sum\limits_{l = 1}^{L}\sum\limits_{i \in \mathcal{P}(l)}\sum\limits_{j \in \mathcal{P}(l), i \neq j} \\ 
&\log \frac{e^{ \left(\mathbf{z}_{i}^{l} \cdot \mathbf{z}_{j}^{l}\right)}}{e^{ \left(\mathbf{z}_{i}^{l} \cdot \mathbf{z}_{j}^{l}\right)} + \sum\limits_{n \in \mathcal{N}(l)} e^{\left(\dot{\mathbf{z}}_{i}^{l} \cdot \mathbf{z}_{n}^{l}\right)}}.
\end{split}
\end{equation}

\paragraph{Cross-scale contrastive learning.} We further associate semantically close moment representations from across different scales. Specifically, we push short-range moment representations closer to semantically close long-range moment representations. This would enable short-range moments to relate to longer video context while long-range features to capture nuanced details of short-range moments. 

As video moment features of layer 0 $\{\mathbf{z}_{j'}^{0}\}$ are the most likely to preserve salient video information compared to other levels, we employ features of the target moments from the lowest level as the anchor set for cross-scale contrastive learning. To construct positive and negative moment set, we utilize features of higher levels $l \in \{1, 2, ..., L\}$ in the feature pyramid corresponding to video moments that involve and do not involve the textual query, respectively. Denoting the set of moment indices in level $l$ that are related to the query as $\mathcal{P}(l)$ and the set of those that are unrelated as $\mathcal{N}(l)$, we define the cross-scale contrastive learning objective as:
\begin{equation}
\begin{split}
\mathcal{L}_{\text{cross}} &= -\sum\limits_{i \in \mathcal{P}(0)}\sum\limits_{l = 1}^{L} \sum\limits_{j \in \mathcal{P}(l)} \\ 
&\log \frac{e^{ \left(\mathbf{z}_{i}^{0} \cdot \mathbf{z}_{j}^{l}\right)}}{e^{ \left(\mathbf{z}_{i}^{0} \cdot \mathbf{z}_{j}^{l}\right)} + \sum\limits_{n \in \mathcal{N}(l)} e^{\left(\mathbf{z}_{i}^{0} \cdot \mathbf{z}_{n}^{l}\right)}}.
\end{split}
\end{equation}

\subsection{Training Objective}
For temporal grounding training, we adopt a focal loss $\mathcal{L}_{\text{cls}}$ for target moment classification and a Distance-IoU loss $\mathcal{L}_{\text{reg}}$ for distance regression from a positive time step $t$ to the target moment. Then, we  combine these losses with our within- and cross-scale contrastive loss:
\begin{equation}
\mathcal{L} = \mathcal{L}_{\text{cls}} + \rho_{\text{reg}} \cdot \mathcal{L}_{\text{reg}} + \rho_{\text{within}} \cdot \mathcal{L}_{\text{within}} + \rho_{\text{cross}} \cdot \mathcal{L}_{\text{cross}},
\end{equation}
where $\rho_{\text{reg}}$, $\rho_{\text{within}}$, and $\rho_{\text{cross}}$ denote hyperparameters to balance the regression, within-scale, and cross-scale contrastive losses, respectively. 

%% file: sections/04_experimental_setup.tex
\section{Experiments}
To validate the effectiveness, we conduct extensive experiments against recent methods for temporal grounding. Subsequently, we perform ablation study to investigate the impact of each component. 

\subsection{Datasets}
Following previous works, we work on five challenging datasets of temporal grounding, which belong to two main categories, \textit{i.e.} 1) Long videos, many queries (Ego4D-NLQ \citep{grauman2022ego4d}, MAD \citep{soldan2022mad}, and TACoS \citep{regneri2013grounding}) and 2) Short videos, few queries (ActivityNet-Captions \citep{krishna2017dense} and Charades-STA \citep{sigurdsson2016hollywood}). 

\noindent\textbf{Ego4D-NLQ} \citep{grauman2022ego4d} consists of egocentric videos recording daily human activities. Each video possesses length from 3.5 to 20 minutes and is associated with 11.6 queries on average. 

\noindent\textbf{MAD} \citep{soldan2022mad} comprises 1.2K hours of movies with 384K queries transcribed from audio description. Since each video is a movie, each exhibits 47 to 202 minutes long.

\noindent\textbf{TACoS} \citep{regneri2013grounding} focuses on cooking topics. The total video length is 10.1 hours and each video is tasked with 143.5 queries for the temporal grounding operation.

\noindent\textbf{ActivityNet-Captions} \citep{krishna2017dense} targets dense video captioning and is subsequently adapted to temporal grounding. Its video length is two minutes on average and the average number of queries per video is approximately 3.65 queries. 

\noindent\textbf{Charades-STA} \citep{sigurdsson2016hollywood} is an action recognition dataset transformed into a temporal grounding one. Each video lasts approximately 30 seconds and possesses 2.4 queries. 

\subsection{Evaluation Metrics}
We report Recall@K at different temporal intersection-over-union $\theta$ (R@K, tIoU = $\theta$) for all datasets. The metric measures the percentage of textual queries whose at least one of the top-$K$ moment predictions temporally overlap with the groundtruth moment more than $\theta$. 

\subsection{Implementation Details}
To fairly compare with previous works and satisfy the scalability of temporal grounding operation for long videos, we adopt video-centric sampling approach \citep{mu2024snag}. For Ego4D-NLQ, we use pre-trained 1) SlowFast video features \citep{feichtenhofer2019slowfast} with BERT textual features \citep{devlin2018bert}, and 2) EgoVLP video and textual features \citep{lin2022egocentric}. For testing, we report R@$\{1,5\}$, tIoU = $\{0.3, 0.5\}$. For MAD dataset, we use CLIP features \citep{radford2021learning} for both videos and texts, and report R@$\{1,5,10,50\}$, tIoU = $\{0.1, 0.3, 0.5\}$. For the TACoS dataset, we use C3D video features \citep{tran2015learning} and GloVe textual features \citep{pennington2014glove}. We report results in terms of R@$\{1,5\}$, tIoU = $\{0.5, 0.7\}$. In addition, we utilize I3D features \citep{carreira2017quo} pre-trained on Kinetics \citep{kay2017kinetics} for Charades-STA and C3D features \citep{tran2015learning} for ActivityNet-Captions experiments. For both datasets, similar to TACoS, we take advantage of GloVe textual features \citep{pennington2014glove}. We report R@$\{1,5\}$, tIoU = $\{0.5, 0.7\}$ for testing on Charades-STA, and R@$\{1,5\}$, tIoU = $\{0.3, 0.5\}$ for testing on ActivityNet-Captions. For more details regarding model architecture, we direct interested readers to the appendix.
For both within-scale and cross-scale contrastive learning implementation, we keep the size of the negative sample set $\mathcal{N}(l)$ in every level $l$ to be equal to the size of the positive video clips $\mathcal{P}(l)$ that correspond to the target video moments. Based upon validation and fair comparison with previous methods, we use $\rho_{\text{ref}} = \rho_{\text{within}} = \rho_{\text{cross}}$ = 1.0.

\subsection{Baselines}
We consider the following temporal grounding models as baselines: 
(i) \textbf{VSL-Net} \citep{zhang2020span} utilizing textual query to highlight regions potential to comprise the target moment; (ii) \textbf{VLG-Net} \citep{soldan2021vlg} modeling temporal grounding as a graph matching problem; (iii) \textbf{Moment-DETR} \citep{lei2021detecting}, a Transformer encoder-decoder architecture that views temporal grounding as a set prediction problem; (iv) \textbf{CONE} \citep{hou2022cone} subsequently slicing a video input into windows, selects relevant windows, and ranks the selected windows to obtain target moments; (v) \textbf{MMN} \citep{wang2022negative}, a Siamese-like network architecture that is trained with video-query and query-video contrastive learning; (vi) \textbf{SSRN} \citep{zhu2023rethinking} enriching anchor frames with additional consecutive frames; (vii) \textbf{G2L} \citep{li2023g2l} measuring moment-query similarities using geodesic distance and quantifies cross-modal interactions with game-theoretic interactions; (viii) \textbf{SOONet} \citep{pan2023scanning}, an anchor-based framework that conducts grounding by pre-ranking, re-ranking, and regression; (ix) \textbf{MESM} \citep{liu2024towards}, a fine-grained moment-query contrastive approach modeled for query word and video moment representations; (x) \textbf{Contrastive-MSAT} \citep{panta2024cross}, applying moment-query contrastive loss supported by a momentum-based memory bank; (xi) \textbf{UVCOM} \citep{xiao2024bridging}, a moment-query contrastive approach for a unified video comprehension framework; (xii) \textbf{SnAG} \citep{mu2024snag} achieving scalable grounding with cross-modal late fusion.

%% file: sections/05_experimental_results.tex
\section{Experimental Results}
\subsection{Main Results}

{\renewcommand{\arraystretch}{1.2}
\begin{table}[t]
\resizebox{\linewidth}{!}{
\begin{tabular}{l|l|cccccc}
\hline
\multicolumn{1}{c|}{\multirow{2}{*}{\textbf{Features}}} & \multicolumn{1}{c|}{\multirow{2}{*}{\textbf{Model}}} & \multicolumn{3}{c}{\textbf{R@1}}           & \multicolumn{3}{c}{\textbf{R@5}}           \\ 
\multicolumn{1}{c|}{}                                   & \multicolumn{1}{c|}{}                                & \textbf{0.3} & \textbf{0.5} & \textbf{Avg} & \textbf{0.3} & \textbf{0.5} & \textbf{Avg} \\ \hline
\multirow{6}{*}{SF+BERT}                                & VSL-Net                                             & 5.45         & 3.12         & 4.29         & 10.74        & 6.63         & 8.69         \\
                                                       & CONE                                                & 10.40        & 5.03         & 7.72         & 22.74        & 11.87        & 17.31        \\
                                                       & SOONet                                              & 8.00         & 3.76         & 5.88         & 22.40        & 11.09        & 16.75        \\
                                                       & SnAG                                                & 9.83         & 6.83         & 8.33         & 27.93        & 19.27        & 23.60        \\
                                                       \rowcolor{Gray}
                                                       & Our model  & \textbf{10.80} & \textbf{7.22} & \textbf{9.49} & \textbf{28.54} & \textbf{20.38} & \textbf{25.06} \\ \hline
\multirow{4}{*}{EgoVLP}                                & VSL-Net                                             & 10.84        & 6.81         & 8.83         & 18.84        & 13.45        & 16.15        \\
                                                       & CONE                                                & 14.15        & 8.18         & 11.17        & 30.33        & 18.02        & 24.18        \\
                                                       & SnAG                                                & 15.72        & 10.78        & 13.25        & 38.39        & 27.44        & 32.92        \\
                                                       \rowcolor{Gray}
                                                       & Ours & \textbf{16.37} & \textbf{11.27} & \textbf{13.96}  &  \textbf{39.97} & \textbf{28.70} &  \textbf{34.43} \\ \hline 
\end{tabular}}
\caption{Results on Ego4D-NLQ.}
\label{tab:ego4d_nlq_results}
\end{table}}

{\renewcommand{\arraystretch}{1.2}
\begin{table*}[t]
\centering
\resizebox{0.75\linewidth}{!}{
\begin{tabular}{l|ccc|ccc|ccc|ccc}
\hline
\multicolumn{1}{c|}{\multirow{2}{*}{\textbf{Model}}} & \multicolumn{3}{c|}{\textbf{R@1}}           & \multicolumn{3}{c|}{\textbf{R@5}}           & \multicolumn{3}{c|}{\textbf{R@10}}          & \multicolumn{3}{c}{\textbf{R@50}}          \\
         & \textbf{0.1} & \textbf{0.3} & \textbf{0.5} & \textbf{0.1} & \textbf{0.3} & \textbf{0.5} & \textbf{0.1} & \textbf{0.3} & \textbf{0.5} & \textbf{0.1} & \textbf{0.3} & \textbf{0.5} \\ \hline
VLG-Net                                             & 3.64         & 2.76         & 1.65         & 11.66        & 9.31         & 5.99         & 17.39        & 14.56        & 9.77         & 39.78        & 34.27        & 24.93        \\
Moment-DETR                                         & 0.31         & 0.24         & 0.16         & 1.52         & 1.14         & 0.28         & 2.79         & 2.06         & 1.20         & 11.08        & 7.97         & 4.71         \\
CONE                                                & 8.90         & 6.87         & 4.10         & 20.51        & 16.11        & 9.59         & 27.20        & 21.53        & 12.82        & 43.36        & 34.73        & 20.56        \\
SOONet                                              & 11.26        & 9.00         & 5.32         & 23.21        & 19.64        & 13.14        & 30.36        & 26.00        & 17.84        & 50.32        & 44.78        & 32.59        \\
SnAG                                                & 10.28        & 8.46         & 5.55         & 24.42        & 20.60        & 13.75        & 32.23        & 27.50        & 19.00        & 52.28        & 46.68        & 35.24        \\
\rowcolor{Gray}
Our model                                           &       \textbf{12.76}       &    \textbf{10.94}          &   \textbf{6.92 }          &     \textbf{26.43}         &     \textbf{22.60}         &     \textbf{15.43}         &     \textbf{34.08}         &     \textbf{29.41}         &   \textbf{20.70}           &       \textbf{54.84}       &        \textbf{48.26}      &      \textbf{37.77}        \\ \hline
\end{tabular}}
\caption{Results on MAD.}
\label{tab:mad_results}
\end{table*}}

{\renewcommand{\arraystretch}{1.2}
\begin{table*}[t]
\centering
\resizebox{0.8\linewidth}{!}{
\begin{tabular}{l|cccc|cccc|cccc}
\hline
\multicolumn{1}{c|}{\multirow{3}{*}{\textbf{Model}}} & \multicolumn{4}{c|}{\textbf{TACoS}}                                  & \multicolumn{4}{c|}{\textbf{ActivityNet-Captions}}                   & \multicolumn{4}{c}{\textbf{Charades-STA}}                           \\  
\multicolumn{1}{c|}{}                                & \multicolumn{2}{c}{\textbf{R@1}} & \multicolumn{2}{c|}{\textbf{R@5}} & \multicolumn{2}{c}{\textbf{R@1}} & \multicolumn{2}{c|}{\textbf{R@5}} & \multicolumn{2}{c}{\textbf{R@1}} & \multicolumn{2}{c}{\textbf{R@5}} \\
                                & \textbf{0.3}    & \textbf{0.5}   & \textbf{0.3}    & \textbf{0.5}   & \textbf{0.5}    & \textbf{0.7}   & \textbf{0.5}    & \textbf{0.7}   & \textbf{0.5}    & \textbf{0.7}   & \textbf{0.5}    & \textbf{0.7}   \\ \hline
VLG-NET                                             & 45.46           & 34.19          & 70.38           & 56.56          & 46.32           & 29.82          & 77.15           & 63.33          & -               & -              & -               & -              \\
MGSL-Net                                            & 42.54           & 32.27          & 63.39           & 50.13          & 51.87           & 31.42          & 82.60           & 66.71          & 63.98           & 41.03          & 93.21           & 63.85          \\
MMN                                                 & 39.24           & 26.17          & 62.03           & 47.39          & 48.59           & 29.26          & 79.50           & 64.76          & -               & -              & -               & -              \\
SSRN                                                & 45.10           & 34.33          & 65.26           & 51.85          & 54.49           & 33.15          & 84.72           & 68.48          & 65.59           & 42.65          & 94.76           & 65.48          \\
G2L                                                 & 42.74           & 30.95          & 65.83           & 49.86          & 51.68           & 33.35          & 81.32           & 67.60          & -               & -              & -               & -              \\
MESM                                                &  52.69            & 39.52         & -           & -          & -           & -         & -           & -          & 61.24            & 38.04          & -          & -          \\
Contrastive-MSAT                                                & 49.77             & 37.99          & 68.31           & 58.31          & 47.73              & 31.21          & 78.06           & 63.63          &   -         &      -     &    -       &    -       \\
UVCOM                                                & 36.39             & 23.32          & -           & -          & -              & -          & -           & -          &   59.25         &      36.64     &    -       &    -       \\
SnAG                                                & 56.44           & 44.86          & 81.15           & 70.66          & 48.55           & 30.56          & 81.71           & 63.41          & 64.62           & 46.26          & 92.55           & 71.94          \\
\rowcolor{Gray}
Ours                                          &  \textbf{58.17}               &      \textbf{47.04}          &         \textbf{84.84}        &         \textbf{73.55}       &       \textbf{54.83}          &    \textbf{33.56}            &       \textbf{84.78}          &         \textbf{68.91}       &         \textbf{66.64}        &      \textbf{47.03}          &      \textbf{93.66}           &    \textbf{72.53}            \\ \hline
\end{tabular}}
\caption{Results on TACoS, ActivityNet-Captions, and Charades-STA.}
\label{tab:tacos_activitynet_charades_results}
\vspace{-10pt}
\end{table*}}

\noindent\textbf{Results on Ego4D-NLQ} (Table \ref{tab:ego4d_nlq_results}). Our framework significantly outperforms recent temporal grounding methods. For example, using SlowFast+BERT features, we outperform previous best method, \textit{i.e.} SnAG, by mean improvements of 1.16\%  and 1.46\% in terms of R@1 and R@5 metrics, respectively. In addition, we accomplish more significant performance gains on the more stringent tIoU threshold of 0.5, denoting more precise moment localization.

\noindent\textbf{Results on MAD} (Table \ref{tab:mad_results}). Similar to results on Ego4D-NLQ, our framework obtains an outstanding improvement over previous temporal grounding methods. Specifically, we enhance SOONet with 1.68 and 2.82 points of R@1 and R@5 on average. Moreover, our model outperforms CONE and SnAG in terms of mean R@1 / R@5 by 3.58 / 6.08 and 2.11 / 1.90 points, respectively, especially for the more stringent tIoU threshold. 

\noindent\textbf{Results on TACoS} (Table \ref{tab:tacos_activitynet_charades_results} (left)). Our model achieves R@1 / R@5 of 47.04\% / 73.55\% at tIoU = 0.5, outperforming the strongest baseline, \textit{i.e.} SnAG, by a substantial margin, \textit{i.e.} +2.18\% R@1 and +2.89\% R@5. Combined with the results on Ego4D-NLQ and MAD, these results demonstrate that our contrastive framework provides beneficial signals to counter informational degradation in the feature pyramid for long-form video grounding. 

\noindent\textbf{Results on ActivityNet-Captions} (Table \ref{tab:tacos_activitynet_charades_results} (middle)). We achieve R@1 / R@5 scores of 33.56\% / 68.91\% at tIoU = 0.7. These results indicate that we outperform SSRN by 0.41\% and 0.43\% with regards to R@1 and R@5, respectively, even though we use the backbone SnAG which is significantly weaker than SSRN.

\noindent\textbf{Results on Charades-STA} (Table \ref{tab:tacos_activitynet_charades_results} (right)). Our model outperforms previous methods by a wide margin. Particularly, we accomplish 47.03\% R@1 and 72.53\% R@5 at tIoU = 0.7, exceeding SSRN by 4.38\% R@1 and 7.04\% R@5. These outcomes on Charades-STA and ActivityNet-Captions show that mutual information signals among video moments contributed by our contrastive framework can polish video moment representations to help temporal grounding on short-form videos.

\subsection{Ablation Study}
\label{sect:ablation}
We conduct extensive experiments on TACoS to study the influence of the design choices.

{\renewcommand{\arraystretch}{1.2}
\begin{table}[t]
\centering
\resizebox{\linewidth}{!}{
\begin{tabular}{l|cc|cc}
\hline
\multicolumn{1}{c|}{\multirow{2}{*}{\textbf{Positive-negative sampling approach}}} & \multicolumn{2}{c|}{\textbf{R@1}} & \multicolumn{2}{c}{\textbf{R@5}} \\
                                              & \textbf{0.3}    & \textbf{0.5}   & \textbf{0.3}    & \textbf{0.5}   \\ \hline
Data augmentation                                                       &    57.00		             &    45.46          &        83.13         &    72.06            \\
Memory bank                                                        &        57.69		       &    46.62            & 84.13                 &         72.94       \\
\rowcolor{Gray}
Ours                                                        & \textbf{58.17}           & \textbf{47.04}          & \textbf{84.84}           & \textbf{73.55}        \\ \hline  
\end{tabular}}
\caption{Ablation results on TACoS with various positive and negative sampling approaches.}
\label{tab:ablation_positive_negative}
\end{table}}

{\renewcommand{\arraystretch}{1.2}
\begin{table}[t]
\centering
\resizebox{0.9\linewidth}{!}{
\begin{tabular}{l|cc|cc}
\hline
\multicolumn{1}{c|}{\multirow{2}{*}{\textbf{Contrastive component}}} & \multicolumn{2}{c|}{\textbf{R@1}} & \multicolumn{2}{c}{\textbf{R@5}} \\
                               & \textbf{0.3}    & \textbf{0.5}   & \textbf{0.3}    & \textbf{0.5}   \\ \hline
w/o within-scale                                     & 57.40           & 46.00          & 83.46           & 72.39          \\
w/o cross-scale                                      & 57.00           & 45.85          & 82.34           & 71.58          \\
\rowcolor{Gray} Ours                                                 & \textbf{58.17}           & \textbf{47.04}          & \textbf{84.84}           & \textbf{73.55}     \\ \hline    
\end{tabular}}
\caption{Ablation results on TACoS with multi-scale contrative components.}
\label{tab:ablation_contrastive_components}
\vspace{-15pt}
\end{table}}

{\renewcommand{\arraystretch}{1.2}
\begin{table}[t]
\centering
\resizebox{0.95\linewidth}{!}{
\begin{tabular}{l|cc|cc}
\hline
\multicolumn{1}{c|}{\multirow{2}{*}{\textbf{Association approach}}} & \multicolumn{2}{c|}{\textbf{R@1}}                      & \multicolumn{2}{c}{\textbf{R@5}}                      \\
& \textbf{0.3}              & \textbf{0.5}              & \textbf{0.3}              & \textbf{0.5}              \\ \hline
Query-query                                                        & 55.61 & 45.06 & 81.25 & 71.75 \\
Moment-query                                                       & 57.00                     & 46.24                     & 82.44                     & 72.37                     \\
CLIP-based moment-moment                                           & 57.13                     & 46.94                     & 83.28                     & 72.96                     \\
\rowcolor{Gray} Ours                                                               & \textbf{58.17}                     & \textbf{47.04}                     & \textbf{84.84}                     & \textbf{73.55}              \\ \hline      
\end{tabular}}
\caption{Ablation results on TACoS with various association approaches.}
\label{tab:ablation_association}
\vspace{-10pt}
\end{table}}


\paragraph{Effect of contrastive components.} We explore what extent each component of our contrastive framework, \textit{i.e.} within- or cross-scale objective, contributes to the overall performance improvement. In Table \ref{tab:ablation_contrastive_components}, cross-scale objective plays a more fundamental role in polishing video moment representations than the within-scale counterpart. Since cross-scale contrastive objective concentrates more upon long-range moment representations by relating them with the short-range moment ones, these results validate our hypothesis that informational degradation is a fundamental problem to resolve in multi-scale temporal grounding.

\paragraph{Effect of moment-moment association.} In addition to our proposed moment-moment association, we experiment with various approaches, \textit{i.e.} moment-query association, query-query association, and one approach to associate video moments but based on the semantic closeness of their corresponding textual queries. For the last approach, we consider two textual queries to be semantically similar if their CLIP-based cosine similarity score is greater than or equal to 0.8 (for positive sampling) and semantically distant if the similarity score is smaller than or equal to 0.2 (for negative sampling). As can be observed in Table \ref{tab:ablation_association}, query-query association performs the worst, as the approach does not polish moment representations. The moment-moment approach outperforms moment-query contrastive learning, but underperforms our method. We hypothesize that there might exist representation conflict between two video moments temporally overlap with each other.

\paragraph{Effect of direct utilization of moment representations.} We study the impact of our direct utilization of moment representations for positive and negative sample generation, and compare with Tube TokenMix \citep{xing2023svformer} as the data augmentation and the momentum-based memory bank approach \citep{panta2024cross}. Table \ref{tab:ablation_positive_negative} shows that we significantly surpass other methods, on average by 1.38 / 1.60 points of R@1 / R@5 over the augmentation approach, and 0.45 / 0.66 points of R@1 / R@5 over the memory bank approach. We hypothesize that while memory bank may maintain a high number of samples for contrastive learning, expensive hyperparameter tuning is essential to achieve an effective performance. 

\subsection{Qualitative Analysis}
In Figure \ref{fig:ego4d_tacos_moment_length_iou}, we observe that our model does not encounter degraded performance when the lengths of the target moments increase. Moreover, we visualize moment predictions of the recent method, \textit{i.e.} SnAG \citep{mu2024snag}, and our model in Figure \ref{fig:temporal_grounding_example}. Even though SnAG could precisely detect the shorter-length moment, it misses the moment of longer length, due to the degraded information issue. In contrast, our framework is able to localize both the short and long moments. We hypothesize that our contrastive framework can hold salient semantics for video moment representations to resolve the degraded signals in the grounding model, thus enhancing the grounding operation towards long video moments.

%% file: sections/06_conclusion.tex
\section{Conclusion}
In this paper, we propose a multi-scale contrastive framework for multi-scale temporal grounding. Essentially, our framework utilizes a query-centric approach to associate temporally separate video moments which correspond to a common textual query to avoid representation conflict. Accordingly, we define a within-scale contrastive objective to model relations among similar-range video moments, and a cross-scsale objective to model relations among cross-range moments. Comprehensive experiments validate the effectiveness of our framework for both short-term and long-term temporal grounding.

%% file: sections/07_acknowledgement.tex
\section*{Acknowledgements}
This research/project is supported by the National Research Foundation, Singapore under its AI Singapore Programme (AISG Award No: AISG3-PhD-2023-08-051T). Thong Nguyen is supported by a Google Ph.D. Fellowship in Natural Language Processing.

%% file: aaai25.bib
@article{zhang2016context,
  title={Context-aware surveillance video summarization},
  author={Zhang, Shu and Zhu, Yingying and Roy-Chowdhury, Amit K},
  journal={IEEE Transactions on Image Processing},
  volume={25},
  number={11},
  pages={5469--5478},
  year={2016},
  publisher={IEEE}
}

@article{burgner2015continuum,
  title={Continuum robots for medical applications: A survey},
  author={Burgner-Kahrs, Jessica and Rucker, D Caleb and Choset, Howie},
  journal={IEEE Transactions on Robotics},
  volume={31},
  number={6},
  pages={1261--1280},
  year={2015},
  publisher={IEEE}
}

@article{claussmann2019review,
  title={A review of motion planning for highway autonomous driving},
  author={Claussmann, Laurene and Revilloud, Marc and Gruyer, Dominique and Glaser, S{\'e}bastien},
  journal={IEEE Transactions on Intelligent Transportation Systems},
  volume={21},
  number={5},
  pages={1826--1848},
  year={2019},
  publisher={IEEE}
}

@article{xu2023boundary,
  title={Boundary-denoising for video activity localization},
  author={Xu, Mengmeng and Soldan, Mattia and Gao, Jialin and Liu, Shuming and P{\'e}rez-R{\'u}a, Juan-Manuel and Ghanem, Bernard},
  journal={arXiv preprint arXiv:2304.02934},
  year={2023}
}

@article{jung2023overcoming,
  title={Overcoming Weak Visual-Textual Alignment for Video Moment Retrieval},
  author={Jung, Minjoon and Jang, Youwon and Choi, Seongho and Kim, Joochan and Kim, Jin-Hwa and Zhang, Byoung-Tak},
  journal={arXiv preprint arXiv:2306.02728},
  year={2023}
}

@inproceedings{zhang2020learning,
  title={Learning 2d temporal adjacent networks for moment localization with natural language},
  author={Zhang, Songyang and Peng, Houwen and Fu, Jianlong and Luo, Jiebo},
  booktitle={Proceedings of the AAAI Conference on Artificial Intelligence},
  volume={34},
  number={07},
  pages={12870--12877},
  year={2020}
}

@inproceedings{zhang2022actionformer,
  title={Actionformer: Localizing moments of actions with transformers},
  author={Zhang, Chen-Lin and Wu, Jianxin and Li, Yin},
  booktitle={European Conference on Computer Vision},
  pages={492--510},
  year={2022},
  organization={Springer}
}

@inproceedings{radford2021learning,
  title={Learning transferable visual models from natural language supervision},
  author={Radford, Alec and Kim, Jong Wook and Hallacy, Chris and Ramesh, Aditya and Goh, Gabriel and Agarwal, Sandhini and Sastry, Girish and Askell, Amanda and Mishkin, Pamela and Clark, Jack and others},
  booktitle={International conference on machine learning},
  pages={8748--8763},
  year={2021},
  organization={PMLR}
}

@article{regneri2013grounding,
  title={Grounding action descriptions in videos},
  author={Regneri, Michaela and Rohrbach, Marcus and Wetzel, Dominikus and Thater, Stefan and Schiele, Bernt and Pinkal, Manfred},
  journal={Transactions of the Association for Computational Linguistics},
  volume={1},
  pages={25--36},
  year={2013},
  publisher={MIT Press One Rogers Street, Cambridge, MA 02142-1209, USA journals-info~…}
}

@inproceedings{grauman2022ego4d,
  title={Ego4d: Around the world in 3,000 hours of egocentric video},
  author={Grauman, Kristen and Westbury, Andrew and Byrne, Eugene and Chavis, Zachary and Furnari, Antonino and Girdhar, Rohit and Hamburger, Jackson and Jiang, Hao and Liu, Miao and Liu, Xingyu and others},
  booktitle={Proceedings of the IEEE/CVF Conference on Computer Vision and Pattern Recognition},
  pages={18995--19012},
  year={2022}
}

@inproceedings{touvron2021going,
  title={Going deeper with image transformers},
  author={Touvron, Hugo and Cord, Matthieu and Sablayrolles, Alexandre and Synnaeve, Gabriel and J{\'e}gou, Herv{\'e}},
  booktitle={Proceedings of the IEEE/CVF international conference on computer vision},
  pages={32--42},
  year={2021}
}

@inproceedings{bodla2017soft,
  title={Soft-NMS--improving object detection with one line of code},
  author={Bodla, Navaneeth and Singh, Bharat and Chellappa, Rama and Davis, Larry S},
  booktitle={Proceedings of the IEEE international conference on computer vision},
  pages={5561--5569},
  year={2017}
}

@inproceedings{soldan2022mad,
  title={Mad: A scalable dataset for language grounding in videos from movie audio descriptions},
  author={Soldan, Mattia and Pardo, Alejandro and Alc{\'a}zar, Juan Le{\'o}n and Caba, Fabian and Zhao, Chen and Giancola, Silvio and Ghanem, Bernard},
  booktitle={Proceedings of the IEEE/CVF Conference on Computer Vision and Pattern Recognition},
  pages={5026--5035},
  year={2022}
}

@inproceedings{sigurdsson2016hollywood,
  title={Hollywood in homes: Crowdsourcing data collection for activity understanding},
  author={Sigurdsson, Gunnar A and Varol, G{\"u}l and Wang, Xiaolong and Farhadi, Ali and Laptev, Ivan and Gupta, Abhinav},
  booktitle={Computer Vision--ECCV 2016: 14th European Conference, Amsterdam, The Netherlands, October 11--14, 2016, Proceedings, Part I 14},
  pages={510--526},
  year={2016},
  organization={Springer}
}

@inproceedings{krishna2017dense,
  title={Dense-captioning events in videos},
  author={Krishna, Ranjay and Hata, Kenji and Ren, Frederic and Fei-Fei, Li and Carlos Niebles, Juan},
  booktitle={Proceedings of the IEEE international conference on computer vision},
  pages={706--715},
  year={2017}
}

@inproceedings{feichtenhofer2019slowfast,
  title={Slowfast networks for video recognition},
  author={Feichtenhofer, Christoph and Fan, Haoqi and Malik, Jitendra and He, Kaiming},
  booktitle={Proceedings of the IEEE/CVF international conference on computer vision},
  pages={6202--6211},
  year={2019}
}

@article{devlin2018bert,
  title={Bert: Pre-training of deep bidirectional transformers for language understanding},
  author={Devlin, Jacob and Chang, Ming-Wei and Lee, Kenton and Toutanova, Kristina},
  journal={arXiv preprint arXiv:1810.04805},
  year={2018}
}

@article{lin2022egocentric,
  title={Egocentric video-language pretraining},
  author={Lin, Kevin Qinghong and Wang, Jinpeng and Soldan, Mattia and Wray, Michael and Yan, Rui and Xu, Eric Z and Gao, Difei and Tu, Rong-Cheng and Zhao, Wenzhe and Kong, Weijie and others},
  journal={Advances in Neural Information Processing Systems},
  volume={35},
  pages={7575--7586},
  year={2022}
}

@inproceedings{tran2015learning,
  title={Learning spatiotemporal features with 3d convolutional networks},
  author={Tran, Du and Bourdev, Lubomir and Fergus, Rob and Torresani, Lorenzo and Paluri, Manohar},
  booktitle={Proceedings of the IEEE international conference on computer vision},
  pages={4489--4497},
  year={2015}
}

@inproceedings{pennington2014glove,
  title={Glove: Global vectors for word representation},
  author={Pennington, Jeffrey and Socher, Richard and Manning, Christopher D},
  booktitle={Proceedings of the 2014 conference on empirical methods in natural language processing (EMNLP)},
  pages={1532--1543},
  year={2014}
}

@inproceedings{carreira2017quo,
  title={Quo vadis, action recognition? a new model and the kinetics dataset},
  author={Carreira, Joao and Zisserman, Andrew},
  booktitle={proceedings of the IEEE Conference on Computer Vision and Pattern Recognition},
  pages={6299--6308},
  year={2017}
}

@article{kay2017kinetics,
  title={The kinetics human action video dataset},
  author={Kay, Will and Carreira, Joao and Simonyan, Karen and Zhang, Brian and Hillier, Chloe and Vijayanarasimhan, Sudheendra and Viola, Fabio and Green, Tim and Back, Trevor and Natsev, Paul and others},
  journal={arXiv preprint arXiv:1705.06950},
  year={2017}
}

@article{zhang2020span,
  title={Span-based localizing network for natural language video localization},
  author={Zhang, Hao and Sun, Aixin and Jing, Wei and Zhou, Joey Tianyi},
  journal={arXiv preprint arXiv:2004.13931},
  year={2020}
}

@article{hou2022cone,
  title={Cone: An efficient coarse-to-fine alignment framework for long video temporal grounding},
  author={Hou, Zhijian and Zhong, Wanjun and Ji, Lei and Gao, Difei and Yan, Kun and Chan, Wing-Kwong and Ngo, Chong-Wah and Shou, Zheng and Duan, Nan},
  journal={arXiv preprint arXiv:2209.10918},
  year={2022}
}

@inproceedings{pan2023scanning,
  title={Scanning only once: An end-to-end framework for fast temporal grounding in long videos},
  author={Pan, Yulin and He, Xiangteng and Gong, Biao and Lv, Yiliang and Shen, Yujun and Peng, Yuxin and Zhao, Deli},
  booktitle={Proceedings of the IEEE/CVF International Conference on Computer Vision},
  pages={13767--13777},
  year={2023}
}

@article{mu2024snag,
  title={SnAG: Scalable and Accurate Video Grounding},
  author={Mu, Fangzhou and Mo, Sicheng and Li, Yin},
  journal={arXiv preprint arXiv:2404.02257},
  year={2024}
}

@inproceedings{wang2022negative,
  title={Negative sample matters: A renaissance of metric learning for temporal grounding},
  author={Wang, Zhenzhi and Wang, Limin and Wu, Tao and Li, Tianhao and Wu, Gangshan},
  booktitle={Proceedings of the AAAI Conference on Artificial Intelligence},
  volume={36},
  number={3},
  pages={2613--2623},
  year={2022}
}

@article{zhu2023rethinking,
  title={Rethinking the video sampling and reasoning strategies for temporal sentence grounding},
  author={Zhu, Jiahao and Liu, Daizong and Zhou, Pan and Di, Xing and Cheng, Yu and Yang, Song and Xu, Wenzheng and Xu, Zichuan and Wan, Yao and Sun, Lichao and others},
  journal={arXiv preprint arXiv:2301.00514},
  year={2023}
}

@inproceedings{li2023g2l,
  title={G2l: Semantically aligned and uniform video grounding via geodesic and game theory},
  author={Li, Hongxiang and Cao, Meng and Cheng, Xuxin and Li, Yaowei and Zhu, Zhihong and Zou, Yuexian},
  booktitle={Proceedings of the IEEE/CVF International Conference on Computer Vision},
  pages={12032--12042},
  year={2023}
}

@inproceedings{guo2020augfpn,
  title={Augfpn: Improving multi-scale feature learning for object detection},
  author={Guo, Chaoxu and Fan, Bin and Zhang, Qian and Xiang, Shiming and Pan, Chunhong},
  booktitle={Proceedings of the IEEE/CVF conference on computer vision and pattern recognition},
  pages={12595--12604},
  year={2020}
}

@inproceedings{yang2023afpn,
  title={AFPN: asymptotic feature pyramid network for object detection},
  author={Yang, Guoyu and Lei, Jie and Zhu, Zhikuan and Cheng, Siyu and Feng, Zunlei and Liang, Ronghua},
  booktitle={2023 IEEE International Conference on Systems, Man, and Cybernetics (SMC)},
  pages={2184--2189},
  year={2023},
  organization={IEEE}
}

@inproceedings{anne2017localizing,
  title={Localizing moments in video with natural language},
  author={Anne Hendricks, Lisa and Wang, Oliver and Shechtman, Eli and Sivic, Josef and Darrell, Trevor and Russell, Bryan},
  booktitle={Proceedings of the IEEE international conference on computer vision},
  pages={5803--5812},
  year={2017}
}

@inproceedings{gao2017tall,
  title={Tall: Temporal activity localization via language query},
  author={Gao, Jiyang and Sun, Chen and Yang, Zhenheng and Nevatia, Ram},
  booktitle={Proceedings of the IEEE international conference on computer vision},
  pages={5267--5275},
  year={2017}
}

@article{liu2021adaptive,
  title={Adaptive proposal generation network for temporal sentence localization in videos},
  author={Liu, Daizong and Qu, Xiaoye and Dong, Jianfeng and Zhou, Pan},
  journal={arXiv preprint arXiv:2109.06398},
  year={2021}
}

@article{xiao2021natural,
  title={Natural language video localization with learnable moment proposals},
  author={Xiao, Shaoning and Chen, Long and Shao, Jian and Zhuang, Yueting and Xiao, Jun},
  journal={arXiv preprint arXiv:2109.10678},
  year={2021}
}

@inproceedings{xiao2021boundary,
  title={Boundary proposal network for two-stage natural language video localization},
  author={Xiao, Shaoning and Chen, Long and Zhang, Songyang and Ji, Wei and Shao, Jian and Ye, Lu and Xiao, Jun},
  booktitle={Proceedings of the AAAI Conference on Artificial Intelligence},
  volume={35},
  number={4},
  pages={2986--2994},
  year={2021}
}

@article{gao2021relation,
  title={Relation-aware video reading comprehension for temporal language grounding},
  author={Gao, Jialin and Sun, Xin and Xu, Mengmeng and Zhou, Xi and Ghanem, Bernard},
  journal={arXiv preprint arXiv:2110.05717},
  year={2021}
}

@inproceedings{soldan2021vlg,
  title={Vlg-net: Video-language graph matching network for video grounding},
  author={Soldan, Mattia and Xu, Mengmeng and Qu, Sisi and Tegner, Jesper and Ghanem, Bernard},
  booktitle={Proceedings of the IEEE/CVF International Conference on Computer Vision},
  pages={3224--3234},
  year={2021}
}

@inproceedings{wang2021structured,
  title={Structured multi-level interaction network for video moment localization via language query},
  author={Wang, Hao and Zha, Zheng-Jun and Li, Liang and Liu, Dong and Luo, Jiebo},
  booktitle={Proceedings of the IEEE/CVF Conference on Computer Vision and Pattern Recognition},
  pages={7026--7035},
  year={2021}
}

@inproceedings{zhang2021multi,
  title={Multi-stage aggregated transformer network for temporal language localization in videos},
  author={Zhang, Mingxing and Yang, Yang and Chen, Xinghan and Ji, Yanli and Xu, Xing and Li, Jingjing and Shen, Heng Tao},
  booktitle={Proceedings of the IEEE/CVF Conference on Computer Vision and Pattern Recognition},
  pages={12669--12678},
  year={2021}
}

@inproceedings{li2022compositional,
  title={Compositional temporal grounding with structured variational cross-graph correspondence learning},
  author={Li, Juncheng and Xie, Junlin and Qian, Long and Zhu, Linchao and Tang, Siliang and Wu, Fei and Yang, Yi and Zhuang, Yueting and Wang, Xin Eric},
  booktitle={Proceedings of the IEEE/CVF Conference on Computer Vision and Pattern Recognition},
  pages={3032--3041},
  year={2022}
}

@inproceedings{li2021proposal,
  title={Proposal-free video grounding with contextual pyramid network},
  author={Li, Kun and Guo, Dan and Wang, Meng},
  booktitle={Proceedings of the AAAI Conference on Artificial Intelligence},
  volume={35},
  number={3},
  pages={1902--1910},
  year={2021}
}

@inproceedings{zhou2021embracing,
  title={Embracing uncertainty: Decoupling and de-bias for robust temporal grounding},
  author={Zhou, Hao and Zhang, Chongyang and Luo, Yan and Chen, Yanjun and Hu, Chuanping},
  booktitle={Proceedings of the IEEE/CVF Conference on Computer Vision and Pattern Recognition},
  pages={8445--8454},
  year={2021}
}

@article{lei2021detecting,
  title={Detecting moments and highlights in videos via natural language queries},
  author={Lei, Jie and Berg, Tamara L and Bansal, Mohit},
  journal={Advances in Neural Information Processing Systems},
  volume={34},
  pages={11846--11858},
  year={2021}
}

@inproceedings{lin2023univtg,
  title={Univtg: Towards unified video-language temporal grounding},
  author={Lin, Kevin Qinghong and Zhang, Pengchuan and Chen, Joya and Pramanick, Shraman and Gao, Difei and Wang, Alex Jinpeng and Yan, Rui and Shou, Mike Zheng},
  booktitle={Proceedings of the IEEE/CVF International Conference on Computer Vision},
  pages={2794--2804},
  year={2023}
}

@article{woo2022explore,
  title={Explore and match: End-to-end video grounding with transformer},
  author={Woo, Sangmin and Park, Jinyoung and Koo, Inyong and Lee, Sumin and Jeong, Minki and Kim, Changick},
  journal={arXiv preprint arXiv:2201.10168},
  volume={1},
  number={4},
  year={2022}
}

@inproceedings{fang2023you,
  title={You can ground earlier than see: An effective and efficient pipeline for temporal sentence grounding in compressed videos},
  author={Fang, Xiang and Liu, Daizong and Zhou, Pan and Nan, Guoshun},
  booktitle={Proceedings of the IEEE/CVF Conference on Computer Vision and Pattern Recognition},
  pages={2448--2460},
  year={2023}
}

@inproceedings{liu2022memory,
  title={Memory-guided semantic learning network for temporal sentence grounding},
  author={Liu, Daizong and Qu, Xiaoye and Di, Xing and Cheng, Yu and Xu, Zichuan and Zhou, Pan},
  booktitle={Proceedings of the AAAI Conference on Artificial Intelligence},
  volume={36},
  number={2},
  pages={1665--1673},
  year={2022}
}

@inproceedings{wang2021exploring,
  title={Exploring cross-image pixel contrast for semantic segmentation},
  author={Wang, Wenguan and Zhou, Tianfei and Yu, Fisher and Dai, Jifeng and Konukoglu, Ender and Van Gool, Luc},
  booktitle={Proceedings of the IEEE/CVF international conference on computer vision},
  pages={7303--7313},
  year={2021}
}

@inproceedings{hu2021region,
  title={Region-aware contrastive learning for semantic segmentation},
  author={Hu, Hanzhe and Cui, Jinshi and Wang, Liwei},
  booktitle={Proceedings of the IEEE/CVF International Conference on Computer Vision},
  pages={16291--16301},
  year={2021}
}

@article{hjelm2018learning,
  title={Learning deep representations by mutual information estimation and maximization},
  author={Hjelm, R Devon and Fedorov, Alex and Lavoie-Marchildon, Samuel and Grewal, Karan and Bachman, Phil and Trischler, Adam and Bengio, Yoshua},
  journal={arXiv preprint arXiv:1808.06670},
  year={2018}
}

@article{zhang2020unsupervised,
  title={An unsupervised sentence embedding method by mutual information maximization},
  author={Zhang, Yan and He, Ruidan and Liu, Zuozhu and Lim, Kwan Hui and Bing, Lidong},
  journal={arXiv preprint arXiv:2009.12061},
  year={2020}
}

@article{bachman2019learning,
  title={Learning representations by maximizing mutual information across views},
  author={Bachman, Philip and Hjelm, R Devon and Buchwalter, William},
  journal={Advances in neural information processing systems},
  volume={32},
  year={2019}
}

@article{chaitanya2020contrastive,
  title={Contrastive learning of global and local features for medical image segmentation with limited annotations},
  author={Chaitanya, Krishna and Erdil, Ertunc and Karani, Neerav and Konukoglu, Ender},
  journal={Advances in neural information processing systems},
  volume={33},
  pages={12546--12558},
  year={2020}
}

@inproceedings{panta2024cross,
  title={Cross-modal Contrastive Learning with Asymmetric Co-attention Network for Video Moment Retrieval},
  author={Panta, Love and Shrestha, Prashant and Sapkota, Brabeem and Bhattarai, Amrita and Manandhar, Suresh and Sah, Anand Kumar},
  booktitle={Proceedings of the IEEE/CVF Winter Conference on Applications of Computer Vision},
  pages={607--614},
  year={2024}
}

@inproceedings{xiao2024bridging,
  title={Bridging the gap: A unified video comprehension framework for moment retrieval and highlight detection},
  author={Xiao, Yicheng and Luo, Zhuoyan and Liu, Yong and Ma, Yue and Bian, Hengwei and Ji, Yatai and Yang, Yujiu and Li, Xiu},
  booktitle={Proceedings of the IEEE/CVF Conference on Computer Vision and Pattern Recognition},
  pages={18709--18719},
  year={2024}
}

@inproceedings{ji2024weakly,
  title={Weakly Supervised Video Moment Retrieval via Location-irrelevant Proposal Learning},
  author={Ji, Wei and Shi, Ruiqi and Wei, Yinwei and Zhao, Shanshan and Zimmermann, Roger},
  booktitle={Companion Proceedings of the ACM on Web Conference 2024},
  pages={1595--1603},
  year={2024}
}

@inproceedings{liu2024towards,
  title={Towards balanced alignment: Modal-enhanced semantic modeling for video moment retrieval},
  author={Liu, Zhihang and Li, Jun and Xie, Hongtao and Li, Pandeng and Ge, Jiannan and Liu, Sun-Ao and Jin, Guoqing},
  booktitle={Proceedings of the AAAI Conference on Artificial Intelligence},
  volume={38},
  number={4},
  pages={3855--3863},
  year={2024}
}

@article{an2023unicom,
  title={Unicom: Universal and compact representation learning for image retrieval},
  author={An, Xiang and Deng, Jiankang and Yang, Kaicheng and Li, Jaiwei and Feng, Ziyong and Guo, Jia and Yang, Jing and Liu, Tongliang},
  journal={arXiv preprint arXiv:2304.05884},
  year={2023}
}

@article{kim2022exploring,
  title={Exploring temporally dynamic data augmentation for video recognition},
  author={Kim, Taeoh and Kim, Jinhyung and Shim, Minho and Yun, Sangdoo and Kang, Myunggu and Wee, Dongyoon and Lee, Sangyoun},
  journal={arXiv preprint arXiv:2206.15015},
  year={2022}
}

@inproceedings{xing2023svformer,
  title={Svformer: Semi-supervised video transformer for action recognition},
  author={Xing, Zhen and Dai, Qi and Hu, Han and Chen, Jingjing and Wu, Zuxuan and Jiang, Yu-Gang},
  booktitle={Proceedings of the IEEE/CVF conference on computer vision and pattern recognition},
  pages={18816--18826},
  year={2023}
}

@article{han2023momentum,
  title={Momentum cross-modal contrastive learning for video moment retrieval},
  author={Han, De and Cheng, Xing and Guo, Nan and Ye, Xiaochun and Rainer, Benjamin and Priller, Peter},
  journal={IEEE Transactions on Circuits and Systems for Video Technology},
  year={2023},
  publisher={IEEE}
}

@inproceedings{nguyen2025meta,
  title={Meta-optimized Angular Margin Contrastive Framework for Video-Language Representation Learning},
  author={Nguyen, Thong and Bin, Yi and Wu, Xiaobao and Dong, Xinshuai and Hu, Zhiyuan and Le, Khoi and Nguyen, Cong-Duy and Ng, See-Kiong and Tuan, Luu Anh},
  booktitle={European Conference on Computer Vision},
  pages={77--98},
  year={2025},
  organization={Springer}
}

@article{nguyen2023demaformer,
  title={Demaformer: Damped exponential moving average transformer with energy-based modeling for temporal language grounding},
  author={Nguyen, Thong and Wu, Xiaobao and Dong, Xinshuai and Nguyen, Cong-Duy and Ng, See-Kiong and Tuan, Luu Anh},
  journal={arXiv preprint arXiv:2312.02549},
  year={2023}
}

@article{nguyen2021contrastive,
  title={Contrastive learning for neural topic model},
  author={Nguyen, Thong and Luu, Anh Tuan},
  journal={Advances in neural information processing systems},
  volume={34},
  pages={11974--11986},
  year={2021}
}

@article{nguyen2022adaptive,
  title={Adaptive contrastive learning on multimodal transformer for review helpfulness predictions},
  author={Nguyen, Thong and Wu, Xiaobao and Luu, Anh-Tuan and Nguyen, Cong-Duy and Hai, Zhen and Bing, Lidong},
  journal={arXiv preprint arXiv:2211.03524},
  year={2022}
}

@article{nguyen2024topic,
  title={Topic Modeling as Multi-Objective Contrastive Optimization},
  author={Nguyen, Thong and Wu, Xiaobao and Dong, Xinshuai and Nguyen, Cong-Duy T and Ng, See-Kiong and Luu, Anh Tuan},
  journal={arXiv preprint arXiv:2402.07577},
  year={2024}
}

@article{nguyen2023improving,
  title={Improving multimodal sentiment analysis: Supervised angular margin-based contrastive learning for enhanced fusion representation},
  author={Nguyen, Cong-Duy and Nguyen, Thong and Vu, Duc Anh and Tuan, Luu Anh},
  journal={arXiv preprint arXiv:2312.02227},
  year={2023}
}

@inproceedings{wu2023infoctm,
  title     = {Infoctm: A mutual information maximization perspective of cross-lingual topic modeling},
  author    = {Wu, Xiaobao and Dong, Xinshuai and Nguyen, Thong and Liu, Chaoqun and Pan, Liang-Ming and Luu, Anh Tuan},
  year      = 2023,
  booktitle = {Proceedings of the AAAI Conference on Artificial Intelligence},
  volume    = 37,
  pages     = {13763--13771},
  url       = {https://arxiv.org/abs/2304.03544}
}

@inproceedings{wu2024modeling,
    title = "Modeling Dynamic Topics in Chain-Free Fashion by Evolution-Tracking Contrastive Learning and Unassociated Word Exclusion",
    author = "Wu, Xiaobao  and Dong, Xinshuai  and Pan, Liangming  and Nguyen, Thong  and Luu, Anh Tuan",
    editor = "Ku, Lun-Wei  and Martins, Andre  and Srikumar, Vivek",
    booktitle = "Findings of the Association for Computational Linguistics ACL 2024",
    month = aug,
    year = "2024",
    address = "Bangkok, Thailand and virtual meeting",
    publisher = "Association for Computational Linguistics",
    url = "https://aclanthology.org/2024.findings-acl.183",
    pages = "3088--3105"
}

@article{nguyen2024kdmcse,
  title={Kdmcse: Knowledge distillation multimodal sentence embeddings with adaptive angular margin contrastive learning},
  author={Nguyen, Cong-Duy and Nguyen, Thong and Wu, Xiaobao and Luu, Anh Tuan},
  journal={arXiv preprint arXiv:2403.17486},
  year={2024}
}
